\documentclass[11pt]{article} %

\usepackage[preprint]{aaai2027} %

\usepackage[hyphens]{url}
\usepackage{graphicx}
\usepackage{xcolor}

\usepackage{natbib}
\usepackage{caption}

\usepackage{amsmath,amssymb,amsfonts}
\usepackage{amsthm}

\usepackage{booktabs}
\usepackage{multirow}

\usepackage{algorithm}
\usepackage{algpseudocode}

\usepackage{listings}

\newcommand{\method}{AWMP}

\title{LLM-Based Exogenous World Modeling for Dynamic Scheduling}
\author{
Ni Zhang\textsuperscript{\rm 1},
Haofeng Yuan\textsuperscript{\rm 2},
Jianan Zhou\textsuperscript{\rm 3},
Shiji Song\textsuperscript{\rm 2},
Zhiguang Cao\textsuperscript{\rm 1}
}

\affiliations{
\textsuperscript{\rm 1}Singapore Management University\quad
\textsuperscript{\rm 2}Tsinghua University\quad 
\textsuperscript{\rm 3}Nanyang Technological University\\
\vspace{1mm}
ni.zhang.2025@phdcs.smu.edu.sg,\quad yhf22@mails.tsinghua.edu.cn\\
jianan004@e.ntu.edu.sg,\quad 
shijis@mail.tsinghua.edu.cn,\quad
zgcao@smu.edu.sg
}

\begin{document}
\maketitle

\begin{abstract}
Dynamic scheduling requires dispatching online arrivals while anticipating future jobs, as poor decisions can compound downstream costs under nonlinear tardiness objectives. We show that a pretrained large language model (LLM) can serve as the exogenous component of a structured world model, forecasting uncertain future job arrivals. Specifically, we exploit an \emph{afterstate decomposition}, which separates known deterministic shop transitions from uncertain exogenous job arrivals, enabling the LLM to forecast only the latter through a compact belief representation. Building on this, we develop \emph{Afterstate World-Model Planning (\method{})}, which combines an evolved, belief-conditioned dispatching strategy with short scenario-based rollouts. A surprise-triggered refresh mechanism queries the LLM only when newly observed arrivals deviate sufficiently from previous forecast, and online residual self-calibration further conditions the LLM on its recent forecast errors to handle nonstationary arrivals. Across diverse arrival distributions, \method{} reduces weighted tardiness by 6-21\% on dynamic job-shop and 22-46\% on flexible job-shop benchmarks, relative to the best of eight classical priority-dispatching rules. 
\end{abstract}

\section{Introduction}
\begin{figure*}[t]
    \centering
    \includegraphics[
        width=\textwidth,
        trim={0.7cm 9.8cm 0.3cm 0.0cm},
        clip
    ]{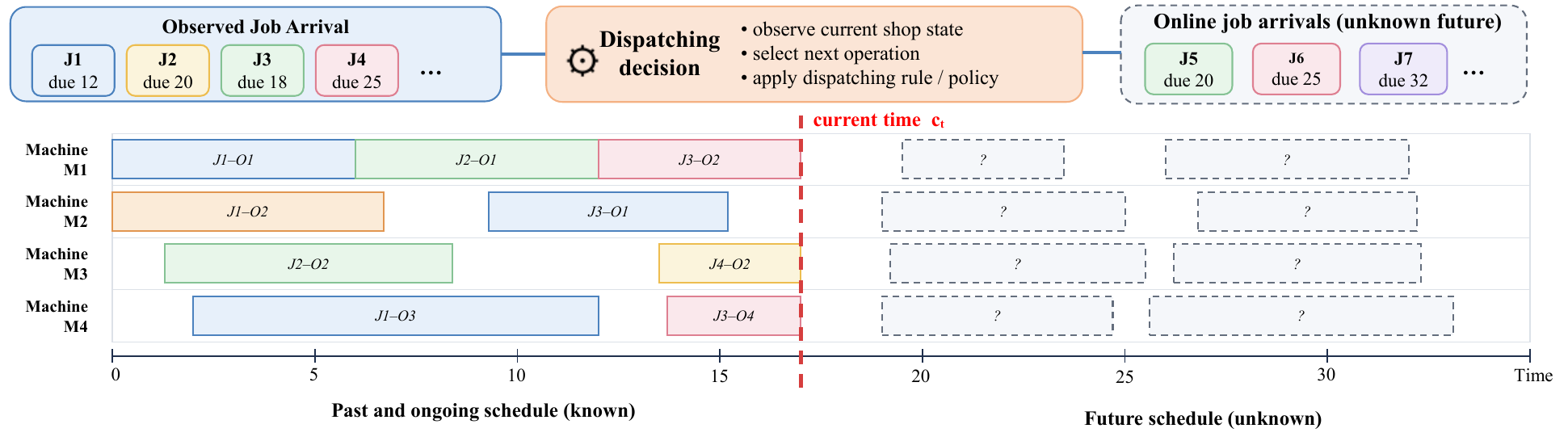}
    \caption{
Illustration of dynamic scheduling with online job arrivals and dispatching decisions under unknown future arrivals.
At the current time $c_t$, the scheduler observes the released jobs
and current shop state and selects the next operation to dispatch.
Here, $J_i$ denotes job $i$, $J_i$--$O_j$ denotes the $j$-th operation of job $i$,
and ``due'' indicates the corresponding job due date.
Future job arrivals remain unknown.
}
    \label{fig:dynamic_scheduling}
\end{figure*}
Scheduling determines how a collection of jobs should be processed on limited machine resources while satisfying operational constraints such as precedence, machine capacity, and delivery commitments \cite{pinedo1992scheduling,smit2025graph}. Classical scheduling formulations assume that all jobs and their processing requirements are known in advance, allowing optimization over a fixed problem instance. 
Real manufacturing systems, however, are inherently dynamic \cite{vieira2003rescheduling, ouelhadj2009survey}. New jobs arrive dynamically over time while machines are already processing previously released work, and the scheduler must repeatedly dispatch operations without knowing the future workload.

Under such uncertainty, scheduling performance is typically evaluated by due-date-oriented objectives. A widely adopted objective is total weighted tardiness, which incurs no cost before a job's due date but imposes a growing penalty once the due date is missed. 
This nonlinear objective creates sharp due-date thresholds: a dispatching decision that appears locally beneficial may occupy critical machine capacity, delay a later high-priority job past its due date, and incur a disproportionate downstream penalty.
Consequently, effective dynamic scheduling requires more than reacting to the currently available operations, but also anticipating \emph{how future job arrivals} will reshape shop congestion and incorporating those expectations into decisions.

Existing approaches address this challenge either by reacting to the current workload or by explicitly modeling the future. 
1) Classical priority-dispatching rules (PDRs), such as weighted shortest processing time (WSPT) and apparent tardiness cost (ATC), are fast and training-free~\citep{panwalkar1977survey,haupt1989survey}, but they rank only the currently available operations and carry no explicit representation of future arrivals.
2) Anticipatory optimization methods, including rolling-horizon replanning, sample-average approximation (SAA), and multiple-scenario approaches (MSA), evaluate decisions against future scenarios sampled from a prespecified stochastic model through repeated reoptimization~\citep{powell2011adp,kleywegt2002saa,bent2004anticipatory,li2025rolling}. Their performance relies heavily on the fidelity of the assumed model, and the repeated reoptimization can be extremely expensive.
3) Recent neural methods learn dispatching policies from simulated trajectories with stochastic arrivals~\citep{chang2022deep,liu2022drlfjsp,liu2023dynamic,liu2024graph,tang2026randomarrivals}. While capable of capturing complex dynamic patterns, they typically entangle the exogenous arrival process with dispatching decisions and require task- and distribution-specific training, with retraining needed once either changes.
These limitations motivate an anticipatory scheduler that can exploit observed arrival patterns without presupposing a parametric stochastic model, and that adapts without heavy retraining or reoptimization.

In this paper, we leverage a pretrained large language model (LLM) to model the \emph{exogenous component} of a structured world model by forecasting future job arrivals, while explicitly accounting for known scheduling dynamics. Specifically, we employ an \emph{afterstate decomposition} to separate the deterministic effects of dispatch actions from the uncertainty associated with future job arrivals. 
At each decision step, a dispatch action transforms the current state into an \emph{afterstate}, representing the resulting shop configuration before subsequent arrivals are incorporated. This afterstate captures endogenous changes, including machine availability, operation progress, and precedence status, governed by known deterministic dynamics, leaving future job arrivals as the sole source of exogenous uncertainty.
Accordingly, the LLM estimates a compact, decision-relevant \emph{belief} over future arrival pressure, represented by the expected numbers of jobs arriving in the early and late portions of a fixed prediction horizon.
The belief summarizes anticipated arrival intensity without specifying individual job identities, machine requirements, or operation-level dynamics.
This compact interface makes the forecast directly usable by a dispatching planner while avoiding the need to learn a high-dimensional transition model.
Building on this decomposition, we develop AWMP, an \emph{Afterstate World-Model Planning} framework for dynamic scheduling. At each decision step, AWMP maps every feasible dispatching action to its deterministic afterstate, and scores the action using both its afterstate features and the belief-derived features that capture future arrival pressure. 
A scenario generator then converts the early-late belief into concrete arrival scenarios, under which the retained action candidates are evaluated through short rollouts, and the action with the best aggregated evaluation is selected for dispatching.

Moreover, AWMP makes the LLM forecast adaptive and efficient for online planning through three mechanisms. 
First, \emph{online residual self-calibration} feeds recent prediction-outcome pairs and forecast errors into subsequent prompts, enabling the LLM to adapt its arrival forecasts to nonstationary patterns through in-context learning.
Second, \emph{belief amortization} caches the forecast across all candidate evaluations and rollouts within a decision step, while the cached belief is also reused across decision steps, where the LLM is queried only when realized arrivals deviate sufficiently from the current belief. This surprise-triggered refresh reduces the number of LLM calls from one per rollout to a handful per episode.
Third, \emph{meta-evolution} searches over compact, human-readable strategy objects, comprising the scoring and value functions, rollout parameters, and forecast-refresh threshold, to automatically discover an effective belief-conditioned planning strategy. 

\newpage
Our main contributions are as follows:
\begin{itemize}

\item We formalize dynamic scheduling through an afterstate decomposition that separates the uncertain, exogenous job arrivals from known, deterministic scheduling dynamics.

\item We instantiate the exogenous arrival-belief component of the structured world model with an LLM to forecast a compact arrival belief, and make this forecast adaptive and efficient through online residual self-calibration and belief reuse with surprise-triggered refresh.

\item We develop AWMP, a belief-conditioned afterstate planner integrating coarse action screening and short-rollout evaluation under fixed scenario generation, with the planning strategy discovered offline via meta-evolution.

\item Experiments across four arrival distributions show that AWMP discovers strategies that outperform the best of eight classical PDRs, reducing total tardiness by 6-21\% on dynamic job-shop instances and 22-46\% on dynamic flexible job-shop instances, while running in seconds.

\end{itemize}

\section{Related Work}

\textbf{Classical dynamic scheduling methods.} Classical methods for dynamic scheduling broadly fall into reactive dispatching and anticipatory optimization. Reactive PDRs, such as WSPT and ATC, re-rank eligible operations whenever the shop state changes~\citep{panwalkar1977survey,haupt1989survey}. They are fast and training-free, but evaluate only the current observable workload with no explicit representation of future arrivals. Anticipatory methods, including rolling-horizon optimization, SAA, and MSA, instead evaluate decisions against expected or sampled future workloads~\citep{powell2011adp,kleywegt2002saa,bent2004anticipatory,li2025rolling}. Their lookahead quality depends on the assumed arrival model, while repeatedly solving the resulting subproblems can be computationally expensive. AWMP occupies a middle ground: it retains low-latency online dispatching but augments it with a compact arrival forecast inferred by an LLM from the observed arrival history, without requiring the arrival pattern to be specified as a parametric stochastic process.

\textbf{Neural dynamic scheduling methods.} Neural combinatorial optimization has introduced graph and attention encoders combined with policy-gradient or value-based reinforcement learning (RL) for routing and scheduling~\citep{kool2019attention,zhang2020l2d}. Recent methods extend learned dispatching policies to dynamic job-shop scheduling problems (JSSPs) and flexible job-shop scheduling problems (FJSPs) with random job arrivals~\citep{luo2020dynamic,liu2022drlfjsp,liu2023dynamic,smit2025stochastic,yu2026dynamic}. These methods can capture complex interactions among jobs, operations, and machines, but they generally learn from specific simulated trajectories and may require retraining when the arrival process or optimization objective changes. In contrast, AWMP trains no neural policy, but uses a frozen pretrained LLM to estimate aggregate arrival pressure, and converts it into dispatching decisions through a compact planning strategy that is evolved offline and frozen before evaluation.

\textbf{Learned world models for planning.} Model-based RL learns a transition model of the environment to evaluate and select future actions. Representative approaches include MuZero~\citep{schrittwieser2020muzero, antonoglou2022stochastic}, Dreamer~\citep{hafner2021dreamerv2}, and TD-MPC~\citep{hansen2022tdmpc}. These methods are designed for settings in which substantial environment dynamics must be learned from interaction. Dynamic scheduling has a different structure: once an action is specified, the transitions in shop status are deterministic and known. Our afterstate decomposition exploits this structure by retaining exact endogenous dynamics and isolating the future arrival stream as the only exogenous uncertainty. AWMP forms a structured world model from two parts: a low-dimensional predictor of exogenous arrival pressure and the deterministic scheduling simulator.

\textbf{LLMs for optimization and heuristic discovery.} LLMs have recently been used to generate and refine algorithms, heuristics, and environment dynamics simulators~\citep{romera2024funsearch,liu2024eoh,ye2024reevo}. AWMP uses LLMs in two separate roles. During online scheduling, a pretrained LLM instantiates the exogenous component of the structured world model by forecasting aggregate arrival pressure. During offline meta-evolution, a proposer LLM evolves a compact, human-readable planning strategy incorporating the scoring rule, value estimator, and rollout parameters, rather than an independent solver program.
\section{Methodology}
\label{sec:method}

\begin{figure*}[t]
\centering
\hspace*{0.001bp}\includegraphics[width=\textwidth]{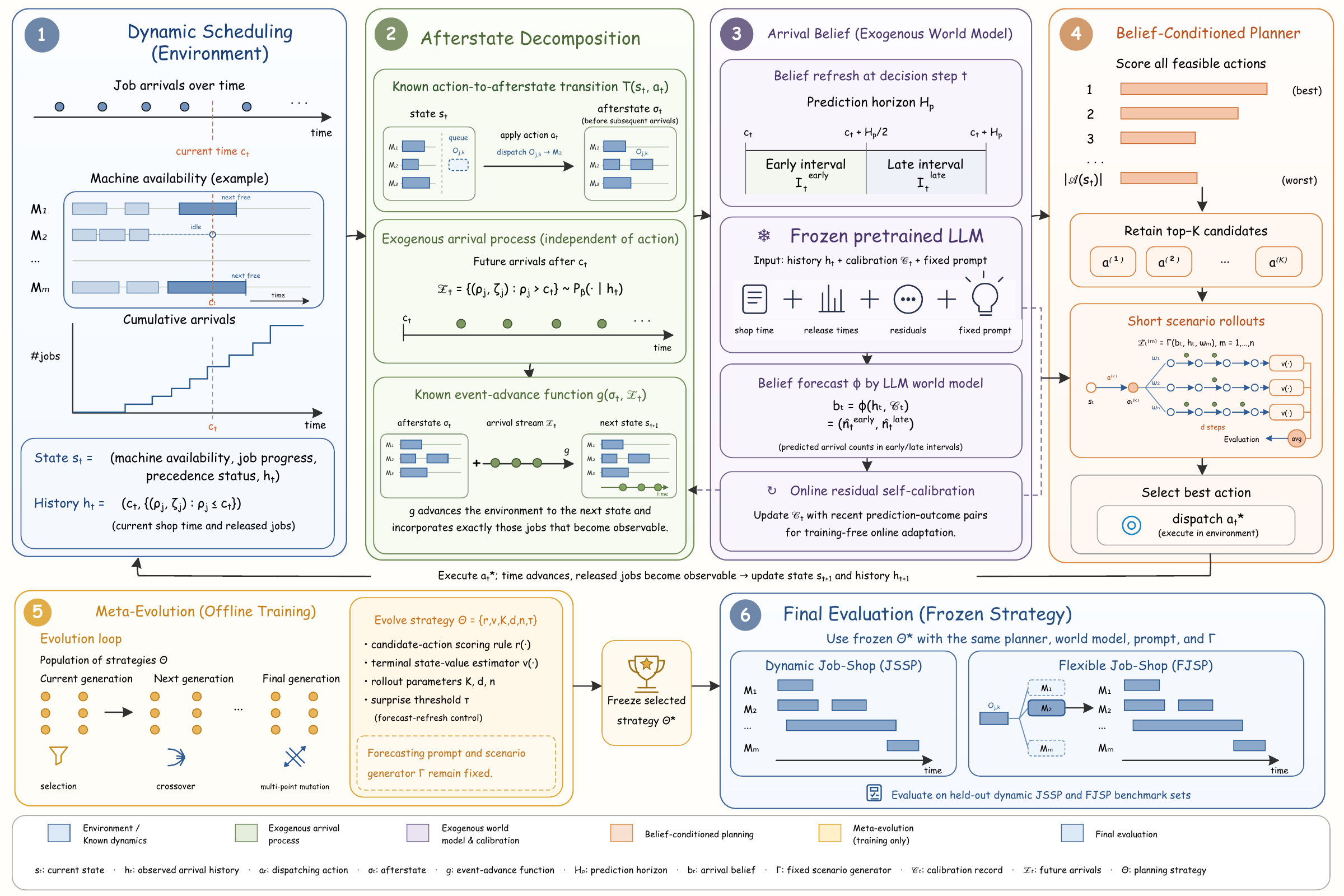}\hspace*{-0.001bp}
\caption{
Overview of our AWMP framework.
1) The dynamic scheduling environment provides the current shop state and the observed arrival history.
2) The afterstate decomposition separates the known, deterministic shop transition from the exogenous job-arrival process.
3) A pretrained LLM forms the exogenous component of the structured world model and, with online residual self-calibration, forecasts a compact early-late arrival belief.
4) The belief-conditioned planner first scores feasible actions, retains the top-$K$ candidates, and evaluates them through short scenario-based rollouts.
5) Meta-evolution optimizes planning strategies offline.
6) The best strategy is frozen and evaluated on held-out dynamic JSSP and FJSP benchmarks.
}
\label{fig:framework}
\end{figure*}

Our methodology is organized around three questions: \textbf{what to forecast}, \textbf{how to forecast}, and \textbf{how to plan} with a structured world model, and AWMP addresses them through a modular design, as illustrated in Fig.~\ref{fig:framework}. First, we formalize dynamic scheduling through an afterstate decomposition, which separates the known endogenous shop dynamics from unknown exogenous job arrivals. This decomposition identifies future arrivals, rather than the complete shop transition, as the only object that requires forecasting (Section~\ref{sec:formulation}). Then, we instantiate the exogenous component with a frozen, pretrained LLM to forecast a low-dimensional belief of early-late future arrival pressure from the observed arrival history, and introduce residual self-calibration to adapt the forecast in an online manner, especially under nonstationary arrivals (Section~\ref{sec:belief}). Building on this, we develop AWMP, a belief-conditioned afterstate planner that first identifies promising actions with a compact scoring rule, and then evaluates them through short rollouts under arrival scenarios generated from the forecast belief (Section~\ref{sec:planner}). Intra- and inter-step belief caching and surprise-triggered belief refresh amortize LLM calls without repeatedly forecasting at every rollout or decision step. 
Finally, meta-evolution discovers a compact, human-readable planning strategy offline, which is then frozen for held-out evaluation (Section~\ref{sec:evolution}).

\subsection{Afterstate Decomposition}
\label{sec:formulation}

We consider dynamic JSSPs and FJSPs, where jobs arrive online and dispatching decisions must be made before observing future workload. Let $t$ index decision steps and $c_t$ denote the corresponding physical shop time. At decision step $t$, the scheduler observes the current shop state $s_t$ and selects a feasible dispatch action $a_t\in\mathcal{A}(s_t)$, where $\mathcal{A}(s_t)$ is the set of all feasible operation-machine assignments. In JSSPs, the machine for each operation is fixed, while in FJSPs, machine selection is part of the action. The state $s_t$ contains the current shop configuration, including machine availability, job progress, precedence status, and observed arrival history $h_t$.
The effect of an action can be separated from that of subsequent arrivals. Applying \(a_t\) first produces an afterstate
\begin{equation}
\sigma_t=T(s_t,a_t),
\label{eq:afterstate_action}
\end{equation}
where $T$ is the known action-to-afterstate transition. The afterstate captures the updated machine availability, operation progress, and, for FJSPs, the selected machine assignment before any new arrivals are incorporated. These changes are \emph{endogenous}: once $s_t$ and $a_t$ are specified, they are fully deterministic.

We define the set of future job arrivals as:
\begin{equation}
\mathcal{Z}_t
=
\left\{
(\rho_j,\zeta_j):\rho_j>c_t
\right\}
\sim
P_\beta(\cdot\mid h_t),
\label{eq:arrival_process}
\end{equation}
where $\rho_j$ is the release time of future job $j$, $\zeta_j$ collects its attributes, such as its processing time, due date, and
weight, and $\beta$ controls the arrival pattern such as intensity.
The arrival process is \emph{exogenous}: dispatching actions do not affect which jobs arrive or their release times.

Nevertheless, an action may change the physical time of the next decision step, and hence which portion of $\mathcal{Z}_t$ becomes observable before that decision. The next state is therefore determined by an event-advance function $g$:
\begin{equation}
s_{t+1}=g(\sigma_t,\mathcal{Z}_t),
\label{eq:afterstate}
\end{equation}
which is likewise known and deterministic given the afterstate $\sigma_t$ and a realization of arrival stream $\mathcal{Z}_t$. The function $g$ advances shop time, executes the deterministic scheduling dynamics, and incorporates exactly the jobs released during the transition. Consequently, $T$ and $g$ need not be learned, and the only unknown component required for exact anticipatory planning is the conditional arrival process $P_\beta(\cdot\mid h_t)$.

This decomposition provides direct insight into how a world model can be structured for dynamic scheduling. Because the endogenous shop transitions are known exactly, AWMP retains them in the scheduling simulator and models the uncertain exogenous arrival process separately. A pretrained LLM estimates the decision-relevant effect of $\mathcal{Z}_t$ as a low-dimensional, compact \emph{belief} from the observed arrival history $h_t$, while the planner combines this belief with the exact transitions $T$ and $g$ for rollout evaluation. Together, the exogenous predictor and deterministic simulator constitute the structured world model used by AWMP.

\subsection{LLM-Based Exogenous World Modeling}
\label{sec:belief}

As discussed above, reliably predicting the complete arrival stream $\mathcal{Z}_t$, such as the exact release times and job attributes, is difficult and unnecessary. We instead forecast a compact projection, referred to as the \emph{belief}, of the future arrival stream $\mathcal{Z}_t$ that reflects the arrival pressure of $P_\beta(\cdot\mid h_t)$ over a temporal horizon relevant to dispatching, without requiring a high-dimensional prediction of individual future jobs. 
We formulate this compact prediction as the exogenous arrival-belief component of the structured world model and instantiate it with a pretrained LLM.

Once a belief forecast is triggered at decision step $t$ (time $c_t$), we divide the prediction horizon $H_p$ into two intervals:
\begin{equation}
I_t^{\mathrm{early}}=[c_t,c_t+H_p/2),\;
I_t^{\mathrm{late}}=[c_t+H_p/2,c_t+H_p].
\end{equation}
The early interval captures near-term arrival pressure that may
immediately affect capacity allocation, while the late interval
provides a coarser indication of subsequent congestion. 
The LLM-based arrival predictor receives the observed arrival history $h_t$ and a residual-calibration record $\mathcal{C}_t$ (introduced below), and returns the belief $b_t$, a forecast of the numbers of job arrivals in the early and late intervals:
\begin{equation}
b_t = \phi(h_t,\mathcal{C}_t) =
(
\widehat{n}_t^{\mathrm{early}}, \widehat{n}_t^{\mathrm{late}}
),
\label{eq:arrival_belief}
\end{equation}
where \(\phi\) denotes the exogenous arrival-belief predictor, and $\widehat{n}_t^{\mathrm{early}}$ and $\widehat{n}_t^{\mathrm{late}}$ are the predicted numbers of arrivals in \(I_t^{\mathrm{early}}\) and \(I_t^{\mathrm{late}}\), respectively. 

The belief $b_t$ provides a shared interface between the arrival predictor and the downstream afterstate planner, and is used in two ways. First, it directly conditions the evolved scoring and value functions, providing side information beyond the currently observable shop status for long-sighted dispatching decisions. Second, a fixed scenario generator, introduced in Section~\ref{sec:planner}, converts this aggregate early--late arrival pressure into concrete rollout scenarios by sampling release times within the corresponding intervals. 
The resulting scenarios allow the planner to evaluate candidate actions through short rollouts while retaining the known deterministic transitions $T$ and $g$ exactly.
The interface of the LLM-based arrival predictor is detailed in Appendix~\ref{app:wm}.

\textbf{Online residual self-calibration.} Observed arrival patterns can drift over the scheduling horizon, making recent forecast errors informative for subsequent predictions. We therefore introduce \emph{online residual self-calibration}, which provides prompt-level closed-loop feedback through recent prediction-outcome residuals.

For a completed previous forecast $b_i$, let $\widehat n_i$ be the predicted number of arrivals over its monitored interval (with ``early'' and ``late'' superscripts omitted for brevity) and $n_i$ the realized arrival count. At decision step \(t\), we maintain a rolling calibration record containing the $L$ most recent forecasts whose prediction horizons have fully elapsed:
\begin{equation}
\mathcal{C}_t
=
\operatorname{recent}_L
\left\{
(n_i, \widehat n_i)
: c_i+H_p\leq c_t
\right\}.
\label{eq:calibration_record}
\end{equation}
When $\mathcal{C}_t$ is nonempty, we summarize the recent forecast bias by the mean signed residual:
\begin{equation}
e_t=
\sum_{(n_i, \widehat n_i)\in\mathcal C_t}
\hspace{-0.8em} (n_i-\widehat n_i) \; / \; |\mathcal C_t|,
\end{equation}
where a positive $e_t$ indicates underestimation, and a negative value indicates overestimation. When $|e_t|$ exceeds a calibration threshold, the subsequent prompt explicitly instructs the LLM-based exogenous predictor to correct this bias. Calibration occurs entirely at the online prompt level: the LLM remains frozen and receives no gradient or parameter update. This mechanism thus provides online adaptation to nonstationary arrivals without fitting a new temporal stochastic model or retraining a neural forecaster network. Section~\ref{sec:calib} shows that the residual self-calibration raises the forecast correlation from $r=0.12$ to $r=0.93$, improving the online arrival belief and downstream scheduling performance.

\subsection{Belief-Conditioned Afterstate Planner}
\label{sec:planner}

The arrival belief provides compact information about future congestion, but it does not by itself determine which action should be dispatched. To this end, AWMP converts the belief $b_t$ into dispatching decisions through a coarse-to-fine planner: a lightweight belief-conditioned scorer first identifies a small set of promising candidate actions, and short scenario-based rollouts then compare and select dispatching actions.
The planner is specified by a compact planning strategy:
\begin{equation}
\Theta = \left\{ r(\cdot),v(\cdot),K,d,n,\tau \right\}, \label{eq:planning_strategy}
\end{equation}
where $r(\cdot)$ is a candidate-action scoring rule, $v(\cdot)$ is a terminal state-value estimator, $K$ is the number of actions retained for rollouts, $d$ is the rollout depth, $n$ is the number of generated scenarios, and $\tau$ is the surprise threshold to trigger belief refresh. Both $r$ and $v$ are compact, human-readable functions rather than neural networks. Their structures, together with the planning parameters, are discovered through meta-evolution introduced in Section~\ref{sec:evolution}.

\textbf{Candidate screening.} For each feasible action $a_t\in\mathcal{A}(s_t)$, AWMP computes a belief-conditioned score from its deterministic afterstate $\sigma_t=T(s_t, a_t)$ and the belief $b_t$, and retains the $K$ best-ranked actions:
\begin{equation}
\mathcal{A}_t^{\operatorname{TopK}} = \operatorname{TopK}_{a_t\in\mathcal{A}(s_t)} r(\sigma_t,b_t).
\end{equation}
This screening can be viewed as a PDR conditioned on the afterstate transition and the arrival belief. It serves as a coarse-grained filter that concentrates the subsequent simulation budget on promising candidates rather than rolling out the complete feasible action set.

\textbf{Scenario generation.} Rollout simulation requires concrete future arrivals, whereas $b_t$ provides only aggregate early–late arrival counts. AWMP bridges this gap through a fixed, non-learned scenario generator. Unlike anticipatory optimization methods such as SAA and MSA, which sample scenarios from a prespecified stochastic model, the generator is fully driven by the LLM-forecast belief inferred from the observed arrival history online:
\begin{equation}
\widetilde{\mathcal{Z}}_t^{(m)} = \Gamma
\left(
b_t,h_t,\omega_m
\right),
\quad
m=1,\ldots,n,
\label{eq:scenario_generator}
\end{equation}
where $\omega_m$ is the random seed for scenario $m$. The generator allocates the forecast arrivals to the corresponding early and late intervals, samples release times within these intervals, and draws the remaining job attributes from a reference distribution fixed before evaluation. Thus, the LLM-forecast belief determines temporal arrival pressure, while $\Gamma$ supplies the remaining details required to instantiate arrival streams for simulation. The scenario generator is trivial, neither learned nor modified during meta-evolution or evaluation.

\textbf{Short-rollout evaluation.} Each retained action $a_t\in\mathcal{A}_t^{\operatorname{TopK}}$ is evaluated under all $n$ generated scenarios. The first simulated dispatch is fixed to $a_t$, and subsequent dispatches are selected by the scoring rule $r$. Given the arrival stream $\widetilde{\mathcal{Z}}_t^{(m)}$, every transition is computed exactly through $T$ and $g$. Note that all rollouts can share the cached root belief $b_t$: once simulated arrivals are realized, their counts can be subtracted from the corresponding intervals to obtain a scenario-specific \emph{residual belief}. This bookkeeping is fully deterministic and requires no additional LLM call.

Let $s_{t+d}^{(m,a_t)}$ denote the state reached after a $d$-step rollout of candidate $a_t$ under scenario $m$. The rollout evaluation of $a_t$ is calculated as the objective cost accumulated during the rollout plus the terminal value estimate $v(s_{t+d}^{(m,a_t)},b_{t,d}^{(m)})$, where $b_{t,d}^{(m)}$ denotes the residual belief at the terminal simulated state. This yields a lightweight $d$-step bootstrap without requiring a long-horizon rollout or repeated optimization. Finally, the action with the best evaluation averaged across the $n$ scenarios is selected for dispatching.

\textbf{Intra- and inter-step belief reuse.} As discussed above, the cached root belief $b_t$ can be shared across all rollouts within a decision step. Reusing it across the $K$ retained actions, $n$ scenarios, and $d$ rollout steps reduces the number of LLM calls from $O(Knd)$ to at most one per decision step.

Moreover, we note that consecutive decision steps often fall within the same prediction horizon, making it unnecessary to query the LLM-based exogenous predictor at every step. AWMP therefore applies a similar belief reuse principle that carries the most recent belief forward by subtracting newly observed arrivals from the corresponding early or late count. A new forecast is triggered only when the discrepancy between the cumulative observed arrivals and those implied by the cached belief exceeds the surprise threshold $\tau$, or when the cached prediction horizon no longer covers the planner's lookahead. Intra-step sharing is exact since all rollouts are conditioned on the same observed history, whereas inter-step reuse is an intentional approximation that trades forecast freshness for fewer LLM calls. Section~\ref{sec:results-calleff} reports the resulting reduction in LLM queries.

\subsection{Meta-Evolution of Planning Strategy}
\label{sec:evolution}

Beyond forecast accuracy, final scheduling performance depends critically on the planning strategy $\Theta$ that translates the forecast belief into specific dispatching decisions. Manually specifying such a strategy is difficult and prone to brittle performance. We therefore formulate the discovery of planning strategies as an LLM-driven meta-evolution loop that automatically evolves strategies to effectively leverage arrival beliefs for producing better dispatching decisions.

Meta-evolution maintains a population of strategies and produces new candidates through selection, crossover, and multi-point mutation, with restart-on-stagnation to preserve diversity. In each iteration, a proposer LLM revises the scoring rule $r(\cdot)$, value estimator $v(\cdot)$, rollout parameters $K,d,n$, and refresh threshold $\tau$ within predefined safety constraints (we force $K \le 6$, $d \le 3$, and $n \le 8$ to avoid unrestricted search) and a structural schema to keep the resulting strategies compact, executable, and human-readable. Candidate fitness is evaluated by weighted tardiness on the training instances, with instance-level costs aggregated in the log domain to reduce domination by a few large-cost instances. Within the meta-evolution loop, the proposer LLM modifies only the strategy object $\Theta$. It neither generates arrival forecasts nor alters the forecasting prompt or scenario generator $\Gamma$. The LLM-based exogenous predictor remains frozen and supplies belief $b_t$ independently during strategy evaluation.

After meta-evolution, the best strategy is selected and frozen. During final evaluation, the scoring rule, value estimator, rollout parameters, refresh threshold, forecasting prompt, scenario generator, and the LLM-based exogenous predictor all remain fixed, with only the residual-calibration record updated online as new arrivals are observed.

\section{Experiments}
\label{sec:experiments}

We evaluate AWMP through four questions: 1) Does AWMP improve scheduling performance across different shop configurations and arrival distributions? 2) Does residual self-calibration improve both forecast accuracy and downstream scheduling? 3) How effectively does belief reuse reduce LLM calls? 4) Can meta-evolution improve the planning strategy and generalize well to held-out instances?

\subsection{Experiment Setup}
\label{sec:setup}

\textbf{Benchmarks.} We construct two dynamic scheduling settings from standard static benchmarks: dynamic JSSPs, based on 80 Taillard instances~\citep{taillard1993}, and dynamic FJSPs, based on 15 Brandimarte and 66 Hurink instances~\citep{brandimarte1993routing, hurink1994tabu}. For each benchmark instance, we preserve its original shop configuration and processing data (e.g., operation precedence, processing times, and machine eligibility), while jobs are released online according to one of four exogenous arrival distributions: pulse train, late-ramp, cyclic, and step-up. These distributions capture periodic bursts, a monotonically increasing arrival rate, cyclic variation, and a mid-horizon rate increase, respectively, as illustrated in Appendix~\ref{app:dists}. We control the arrival intensity $\beta$ to scale the scheduling horizon by a factor of five relative to the base benchmarks to avoid trivial instances. Scheduling performance is measured by total weighted tardiness, where jobs completed before their due dates incur no penalty, while unfinished jobs are charged at the end of the scheduling horizon, as detailed in Appendix~\ref{app:obj}. %

\begin{table*}[t]
\centering
\vspace{2mm}
\caption{Average objective (total weighted tardiness) on dynamic JSSP and FJSP instances under four arrival distributions (lower is better). Best PDR reports the best result among the eight baseline PDRs.}
\label{tab:main}
\setlength{\tabcolsep}{4pt}
\begin{tabular}{c@{\hspace{3.5pt}}c|cccccc@{\hspace{6pt}}|@{\hspace{6pt}}c}
\toprule
Benchmark & Distribution & Myopic & Rolling & MSA & SAA & EV & Best PDR & \method{} (Ours) \\
\midrule
\multirow{4}{*}{\shortstack[c]{Dynamic\\JSSP}}
 & Pulse train & 79931 & 67930 & 66509 & 66568 & 66178 & 32923 & \textbf{29238}{\footnotesize~(-11.2\%)} \\
 & Late-ramp   & 70584 & 59654 & 58544 & 58957 & 58380 & 34841 & \textbf{27619}{\footnotesize~(-20.7\%)} \\
 & Cyclic      & 82708 & 70243 & 69530 & 69561 & 68841 & 34486 & \textbf{32483}{\footnotesize~(-5.8\%)} \\
 & Step-up     & 78631 & 66590 & 65719 & 65885 & 65388 & 32686 & \textbf{29630}{\footnotesize~(-9.4\%)} \\
\midrule
\multirow{4}{*}{\shortstack[c]{Dynamic\\FJSP}}
 & Pulse train & 836  & 369 & 384 & 342 & 315 & 212 & \textbf{163}{\footnotesize~(-23.1\%)} \\
 & Late-ramp   & 1543 & 559 & 912 & 591 & 561 & 456 & \textbf{258}{\footnotesize~(-43.4\%)} \\
 & Cyclic      & 1164 & 383 & 512 & 395 & 372 & 300 & \textbf{161}{\footnotesize~(-46.3\%)} \\
 & Step-up     & 1130 & 386 & 582 & 528 & 497 & 265 & \textbf{207}{\footnotesize~(-21.9\%)} \\
\bottomrule
\end{tabular}
\end{table*}

\textbf{Baselines.} We compare AWMP with two groups of baselines. The first contains eight classical PDRs: WSPT, ATC, shortest processing time (SPT), earliest due date (EDD), most work remaining (MWKR), first-in-first-out (FIFO), minimum slack (MS), and cost over time (COVERT). They re-rank currently feasible operations at every decision step whenever the shop state changes or new job arrivals occur. The second group comprises five optimization-based online schedulers: Myopic, Rolling-horizon, MSA, SAA, and expected-value optimization (EV). Myopic optimizes over currently released jobs only; Rolling-horizon periodically reoptimizes over a moving planning window; MSA selects decisions through multiple-scenario consensus; SAA optimizes a sample-average objective over sampled arrival scenarios; and EV solves a deterministic approximation under expected future arrivals. All five methods use OR-Tools~\citep{perron2011ortools} to solve their subproblems, and are allocated computational budgets at least as large as AWMP. We use Claude Haiku 4 as the frozen LLM for online arrival-belief prediction and Claude Sonnet 4 as the proposer LLM for offline meta-evolution.

\subsection{Overall Performance}
\label{sec:overall}
We evaluate \method{}, i.e., the evolved planning strategy that leverages the belief produced by the LLM-based exogenous predictor, on dynamic JSSPs and FJSPs under four arrival distributions. We report the average objective (total weighted tardiness) over non-trivial instances, defined as those on which at least one of the eight PDRs incurs positive tardiness. Table~\ref{tab:main} shows that \method{} achieves the best objective in all eight benchmark-distribution combinations, reducing total weighted tardiness relative to the best PDR by 5.8\%--20.7\% on dynamic JSSPs and by 21.9\%--46.3\% on dynamic FJSPs, and outperforming all five optimization-based baselines in every setting. These gains require only a handful of LLM queries per episode (Section~\ref{sec:results-calleff}), giving runtimes comparable to or below Rolling-horizon and approximately one order of magnitude below other anticipatory baselines.

Importantly, unlike neural scheduling methods that typically require distribution-specific retraining, AWMP uses a single evolved strategy
that is frozen and evaluated across all four arrival distributions without per-distribution retuning. Its improvements therefore extend
beyond the arrival process encountered during strategy evolution, covering bursty, increasing, cyclic, and step-changing arrival patterns.
This suggests that the combination of history-based arrival beliefs and the evolved afterstate planner generalizes across different forms of
arrival dynamics.

We also observe that the optimization-based baselines do not consistently outperform classical PDRs. In dynamic scheduling, anticipatory performance depends on both optimization strength and how well the assumed future workload reflects the arrivals that are eventually realized. Methods such as
SAA, MSA, and EV optimize decisions with respect to sampled or expected future arrivals; when these assumptions deviate from the realized temporal arrival pattern, stronger optimization may still produce decisions that are suboptimal for the actual future workload. This effect is particularly consequential under weighted tardiness, where an early decision can occupy critical capacity and lead to substantial downstream penalties.

In contrast, classical PDRs make no explicit assumptions about future arrivals and continuously re-rank available operations as the shop state changes, thereby avoiding errors induced by an inaccurate explicit future model. AWMP combines the responsiveness of online dispatching with explicit anticipation by inferring future arrival pressure from the
observed history and incorporating it into short-horizon planning.

\subsection{Effect of Residual Self-Calibration}
\label{sec:calib}

\begin{figure}[t]
\centering
\includegraphics[width=0.9\columnwidth]{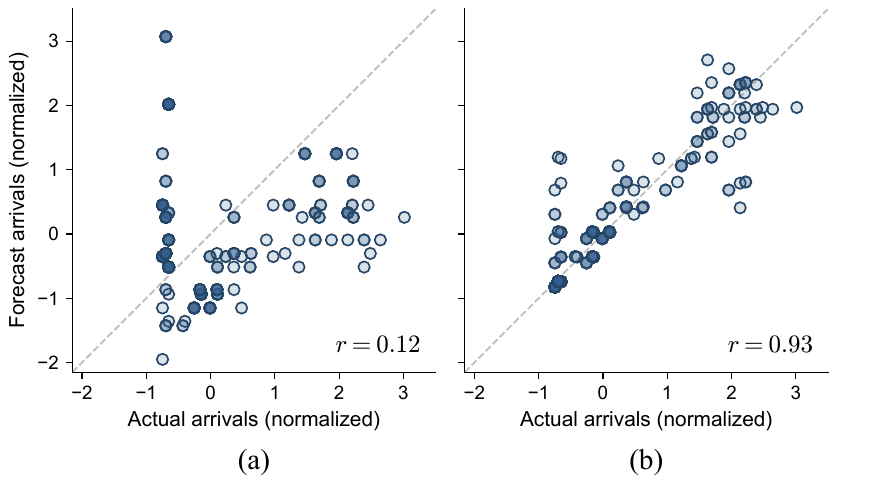}
\caption{Forecast correlation (a) without and (b) with residual self-calibration, increasing from $r=0.12$ to $r=0.93$.}
\label{fig:calib}
\end{figure}

We evaluate the forecast accuracy and the effect of residual self-calibration on dynamic JSSP instances under a sinusoidally drifting arrival intensity. The near-term forecast correlation increases from $r=0.12$ using only the observed arrival history to $r=0.93$ after incorporating recent prediction--outcome residuals (Fig.~\ref{fig:calib}). We compute the latter from 342 prediction--outcome pairs sampled from the same dynamic JSSP instances and arrival distribution, after centering the predictions and outcomes within each instance. This closed-loop calibration enables the arrival belief to more closely track the evolving arrival dynamics without parameter updates.

The improved forecast also leads to better scheduling performance. Under the same planning strategy, replacing the raw belief with the calibrated belief reduces the objective from 6771 to 4011, corresponding to a 41\% reduction. The sampled prediction--outcome pairs are obtained from a separate diagnostic run and are not included in the main evaluation results.

\subsection{LLM Call Efficiency}
\label{sec:results-calleff}

We compare three querying regimes under the same running setting: 
1) naive per-rollout querying, 
2) w/ intra-step belief caching, and 
3) w/ inter-step reuse and surprise-triggered refresh. 
Naive querying requires 7191 LLM calls per episode. 
Intra-step belief caching reduces this count to 393, leading to an 18$\times$ reduction, while incorporating inter-step reuse and surprise-triggered refresh further reduces it to only 18, approximately $400\times$ fewer calls than naive querying. 
The reduction comes from sharing the same root belief across candidate actions and rollout scenarios within a decision step, thereby avoiding repeated forecasts for identical observed histories. 
Across consecutive decision steps, the cached belief is carried forward and refreshed only when realized arrivals deviate sufficiently from the current belief or when the prediction horizon is no longer adequate. 
At the measured latency of approximately 0.84\,s per call, the naive regime requires well over an hour per episode due to the latency of LLM calls, whereas AWMP with inter- and intra-step reuse finishes its queries within seconds. 
These results show that belief reuse substantially amortizes the cost of online LLM forecasting for repeated dispatching decisions, while preserving the same belief-conditioned planning procedure.

\subsection{Meta-evolution}
\label{sec:results-meta}

\begin{figure}[t]
\centering
\includegraphics[width=\columnwidth]{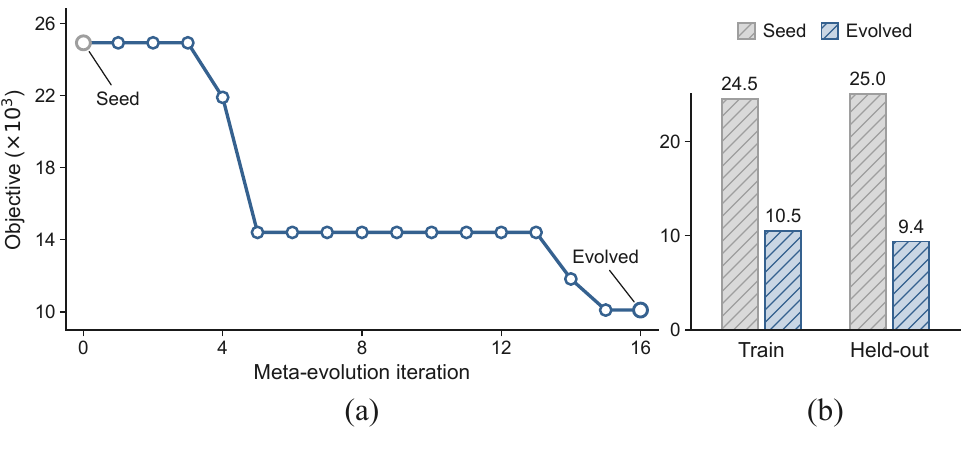}
\caption{(a) Meta-evolution trajectory; (b) seed vs.\ evolved strategy performance on training and held-out instances.}
\label{fig:meta}
\end{figure}

Figure~\ref{fig:meta} tracks the meta-evolution process under a strict train-test split. 
We use TA01 and TA02 exclusively for strategy evolution, where the scoring rule, 
value estimator, rollout parameters, and refresh threshold are iteratively refined 
according to the training objective. After meta-evolution terminates, we select the 
single best-performing strategy, freeze the resulting planning strategy, and evaluate 
it directly on the disjoint held-out instances TA03 and TA04 without any further 
strategy adaptation or instance-specific tuning. As shown in Figure~\ref{fig:meta}, 
the objective decreases substantially over the course of meta-evolution, indicating 
that the search process progressively discovers more effective ways of translating 
the arrival belief into dispatching decisions. Relative to the initial seed strategy, 
the evolved strategy reduces the objective by $57\%$ on the training instances and 
by $62\%$ on the held-out instances. Notably, the improvement on the unseen instances 
is comparable to that observed on the training instances. This suggests that 
meta-evolution does not merely fit the few instances used during strategy search, 
but discovers planning rules that remain effective beyond the instances used for 
evolution. Since the final planning strategy is frozen before evaluation, the gain 
on TA03 and TA04 reflects the transferability of the evolved belief-conditioned 
decision logic rather than further strategy adaptation at test time. Overall, these 
results provide evidence that a compact strategy evolved from only a small number of 
training instances can remain effective on previously unseen dynamic scheduling instances.

\section{Conclusion}

We introduced AWMP, an afterstate world-model planning framework for dynamic scheduling under online arrivals. The afterstate decomposition separates known deterministic shop dynamics from the exogenous uncertain arrival stream, forming a structured world model in which a pretrained LLM models the exogenous component while the scheduling simulator retains the known endogenous transitions exactly. Residual self-calibration adapts the belief forecast online under nonstationary arrivals, while a meta-evolved, belief-conditioned planner converts it into dispatching decisions through candidate screening and short rollouts under a fixed, non-learned scenario generator. Experiments show that combining a low-dimensional LLM-based forecast of exogenous uncertainty with exact scheduling dynamics supports effective anticipatory scheduling without learning the complete shop transition or training a task-specific neural dispatching policy. Future work includes enriching the arrival belief with more job-level information and evaluating AWMP on real-world and more diverse arrival processes. We will also explore more efficient forecasting models to further reduce online inference latency. More broadly, the structured world-model framework may extend to robust optimization and stochastic programming settings.

Our study has several limitations. First, AWMP represents future arrivals through a compact early--late workload belief rather than detailed job-level predictions. This design keeps the forecasting problem low-dimensional and supports stable online calibration, avoiding the difficulty of predicting the complete future arrival stream. However, richer representations that additionally capture job attributes, such as processing routes, priorities, or machine requirements, may provide more informative guidance when future workloads are highly heterogeneous. Second, our experiments construct dynamic instances from standard JSSP and FJSP benchmarks under four controlled arrival patterns. These patterns cover several representative forms of temporal variation and allow controlled evaluation across different arrival dynamics. However, real manufacturing systems may exhibit more diverse arrival behaviors and operational conditions, and evaluation on real scheduling traces would provide a broader assessment of the framework. Finally, AWMP uses a pretrained LLM for online belief updates. Belief reuse and surprise-triggered refresh substantially limit the associated inference overhead. Nevertheless, the runtime of belief updates is still affected by the latency of the underlying model, and more efficient forecasting models could further improve online deployment.
\clearpage
\bibliography{refs}

\clearpage
\newpage
\appendix
\section{Arrival Distributions}
\label{app:dists}

\begin{figure}[!h]
\centering
\includegraphics[width=\columnwidth]{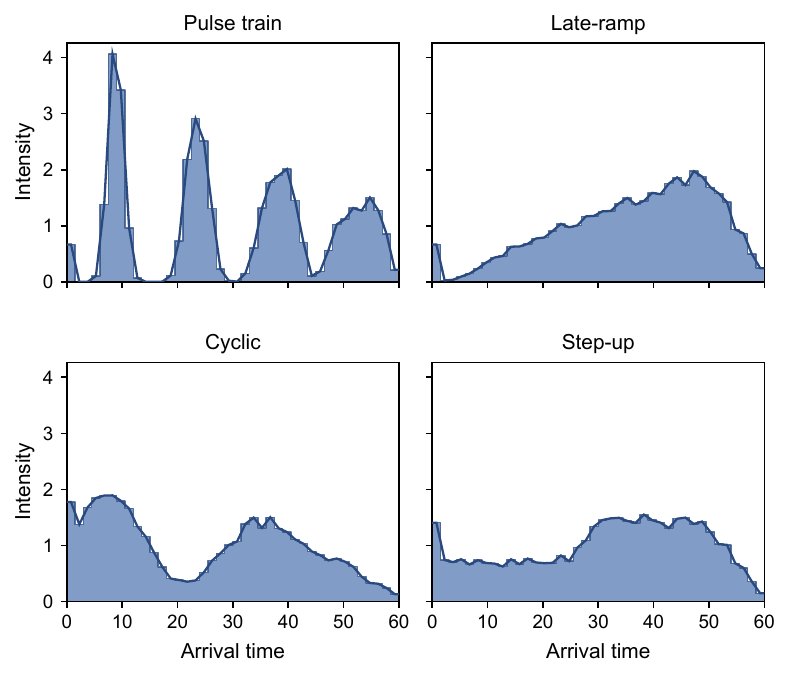}
\caption{Empirical arrival-rate profile of each distribution family over the
arrival window ($200$ seeds, $\beta=1.5$).}
\label{fig:dists}
\end{figure}

We construct each dynamic instance by assigning release times to the jobs of its underlying static benchmark. Job releases are restricted to an arrival window of length $H_{\mathrm{arr}}=0.6H$, where $H$ is the episode horizon. The first job is available at $t=0$, while the release times of the remaining jobs are sampled from one of four distribution families. The intensity parameter $\beta$ scales the peak arrival rate. Each family exhibits a temporal structure that can, in principle, be inferred from the observed arrival history. Fig.~\ref{fig:dists} shows the empirical arrival-intensity profiles aggregated over 200 random seeds.

\begin{itemize}
    \item \textbf{Pulse train.} Arrivals are concentrated around $K_{\text{pulse}}=4$ evenly spaced burst centers, with Gaussian temporal jitter scaled by $0.02H_{\mathrm{arr}}$. This produces a sequence of short, periodic workload surges.
    \item \textbf{Late-ramp.} Arrival times are generated as:
    \begin{equation*}
    t_j=\sqrt{U_j}\,H_{\mathrm{arr}}, \text{ where } U_j\sim\operatorname{Unif}(0,1),
    \end{equation*}
    resulting in an arrival rate that increases monotonically over the window. The expected normalized release time is $2/3$, producing a back-loaded workload.
    \item \textbf{Cyclic.} The arrival intensity follows two sinusoidal cycles over the arrival window:
    \begin{equation*}
    \lambda(t) = \lambda_0 \left(1+0.9\sin\left(4\pi t/H_{\mathrm{arr}}\right)\right).
    \end{equation*}
    This distribution alternates between recurring high- and low-intensity periods.
    \item \textbf{Step-up.} One third of the jobs are released uniformly during the first half of the window, while the remaining two thirds are released during the second half. Consequently, the expected arrival rate doubles at the midpoint.
\end{itemize}

\section{World-Model Interface}
\label{app:wm}

\paragraph{World-model input and output.} Whenever a belief refresh is triggered at decision step $t$, the world model receives the observed arrival history $h_t$, including the current physical time, prediction horizon, number of jobs released so far, and their release-time log. It returns the arrival belief:
\begin{equation*}
    b_t=(\widehat n_t^{\mathrm{early}},\widehat n_t^{\mathrm{late}}),
\end{equation*}
where $\widehat n_t^{\mathrm{early}}$ and $\widehat n_t^{\mathrm{late}}$ are nonnegative integer estimates of the numbers of new jobs arriving in the early and late portions of the prediction horizon, respectively. The LLM predicts only temporally aggregated arrival pressure; it does not predict individual job identities, exact release times, processing routes, or other job attributes. The concrete release times and job attributes required for rollout simulation are instantiated by the fixed scenario generator described in Section~\ref{sec:planner}.

\paragraph{Prompt.} Each world-model query combines a fixed forecasting instruction, the observed arrival history, the residual-calibration record, and a predefined JSON output schema. The model is instructed to return only
\begin{equation*}
\texttt{\{"early": e, "late": l\}},
\end{equation*}
where $e$ and $l$ are the predicted arrival counts. The instruction, output schema, and all other prompt components remain fixed during meta-evolution and final evaluation.

\section{Objective}
\label{app:obj}

The objective is total weighted tardiness:
\begin{equation*}
\operatorname{Cost}
=
\sum_j
w_j\max\left(0,\widetilde C_j-d_j\right),
\end{equation*}
where
\begin{equation*}
d_j=\rho_j+\mu p_j,
\quad
w_j=\frac{p_j}{\bar p}.
\end{equation*}
Here, \(\rho_j\) is the release time of job \(j\), \(p_j\) is its total processing workload, \(\bar p\) is the mean workload across all jobs in the instance, and \(\mu\) controls due-date tightness. For a job completed within the episode, \(\widetilde C_j\) is the completion time of its final operation. An unfinished job is assigned \(\widetilde C_j=H\), where \(H\) is the episode horizon, ensuring that unfinished work contributes a terminal tardiness penalty rather than being omitted from the objective.
The hinge term \(\max(0,\widetilde C_j-d_j)\) creates a due-date threshold: a job incurs no penalty before its due date and a linearly increasing penalty thereafter. The due date accounts for the job’s own processing workload through \(\mu p_j\), while \(w_j=p_j/\bar p\) assigns proportionally greater importance to processing-intensive jobs.

\end{document}